\documentclass{article}

\usepackage[final]{neurips_2021}

\usepackage[utf8]{inputenc} % allow utf-8 input
\usepackage[T1]{fontenc}    % use 8-bit T1 fonts
\usepackage{hyperref}       % hyperlinks
\usepackage{url}            % simple URL typesetting
\usepackage{booktabs}       % professional-quality tables
\usepackage{amsfonts}       % blackboard math symbols
\usepackage{nicefrac}       % compact symbols for 1/2, etc.
\usepackage{microtype}      % microtypography
\usepackage{xcolor}         % colors

\usepackage{caption}
\usepackage{subcaption}
\usepackage{graphicx}
\usepackage{wrapfig}
\usepackage{enumitem}

\usepackage{makecell, caption, booktabs, multirow, siunitx}

\newcommand{\otoprule}{\midrule[\heavyrulewidth]}
\usepackage{lipsum, fancyvrb}

\title{Robust Calibration For Improved Weather Prediction Under Distributional Shift}

\author{%
  Sankalp Gilda\\%\thanks{Use footnote for providing further information about author (webpage, alternative address)---\emph{not} for acknowledging funding agencies.} \\
  ML Collective\\%Department of Computer Science\\
  \texttt{sankalp.gilda@gmail.com} \\
  \And
  Neel Bhandari\\
  Department of Computer Science and Engineering\\
  RV College of Engineering\\
  \texttt{neelbhandari.cs18@rvce.edu.in} \\
  \And
  Wendy Mak\\
  ML Collective\\
  \texttt{wwymak@gmail.com} \\
  \And
  Andrea Panizza\\
  ML Collective\\
  \texttt{andrea.panizza75@gmail.com} \\
}

\begin{document}

\maketitle

\begin{abstract}
 In this paper, we present results on improving out-of-domain weather prediction and uncertainty estimation as part of the \texttt{Shifts Challenge on Robustness and Uncertainty under Real-World Distributional Shift} challenge. We find that by leveraging a mixture of experts in conjunction with an advanced data augmentation technique borrowed from the computer vision domain, in conjunction with robust \textit{post-hoc} calibration of predictive uncertainties, we can potentially achieve more accurate and better-calibrated results with deep neural networks than with boosted tree models for tabular data. We quantify our predictions using several metrics and propose several future lines of inquiry and experimentation to boost performance.
\end{abstract}

\section{Introduction}\label{sec:introduction}
%Machine learning models are often trained with the "closed-world" assumption; models learn to minimize losses on the training data with the intended aim to deliver high performance on an independent testing data. Overparameterized models may have exceptional performance on testing sets, but then fail on out-of-distribution or "unconventional" examples [1-3]. Failure modes point to problems with over-reliance on spurious correlations [4, 5] or exploitation of inappropriate biases used in inference [6, 7].  For example, some image classifiers have been shown to rely on the background details instead of the actual object of interest for prediction [5, 8]. Problematically as well, deep neural networks(DNN) have been shown to make overconfident predictions on inputs far away from training data [9]. %\cite{Gal2016Uncertainty; Ref9}. 

Machine learning is increasingly being applied across several domains, finding ever new applications in the real world, in fields ranging from astrophysics \citep{gilda_deepremap_abstract_aas, gilda_deepremap, gilda_beyond_mirkwood, gilda_gamma_ray, gilda_mirkwood, gilda_mirkwood_aasabstract, gilda_mirkwood_code, gilda_mirkwood_hstproposal, gilda_feature_selection}, to recommendations systems \citep{recsys_aher2013combination, recsys_doshi2018agroconsultant, recsys_portugal2018use}, to drug discovery \citep{drug_discovery_lavecchia2015machine, drug_discovery_vamathevan2019applications, drug_discovery_zhang2017machine}. A key assumption held by several of these models is that training and test data are independent and identically distributed (IID); this assumption rarely holds up to scrutiny in the real-world, where the models must process unseen and unpredictable distributions. This leads to several machine learning models providing unsatisfactory performance in production. The development of models that are robust to distributional shifts, therefore, is an important goal to work towards.

A prime exemplar for a field where distributional shifts are abundant over time is weather prediction. Weather prediction requires that models provide consistently satisfactory performance across both as a function of space (latitude, longitude, climate) and time (of day, month, year). Due to geographical and population constraints, the distribution of weather recording services is non-uniform in nature. 
While there have been several works in the field of weather prediction, the \texttt{Shifts Challenge on Robustness and Uncertainty under Real-World Distributional Shift} presents a unique opportunity to develop and apply models that are robust to such effects and also yield sensible uncertainty estimates. We present here an empirical study on the effects of regularization, calibration, and data augmentation in a multi-domain training environment.

Our main contributions are thus. First, we leverage a mixture density network \citep[MDN,][]{mdn0, mdn1, gilda_cfht, gilda_cfht_neurips} with $\beta-$likelihoods and demonstrate competitive performance relative to two commonly used boosted-tree models, \texttt{NGBoost} \citep{ngboost} and \texttt{CatBoost} \citep{catboost}. Second, we demonstrate the necessity of regularization while training the MDN. Specifically, we use moment exchange \citep[MoEx,][]{moex}, a data augmentation method originally developed for use in computer vision (CV), to regularize our MDN successfully. Third, we demonstrate the necessity of calibrating predictions, and utilize a state-of-the-art \textit{post-hoc} calibration method \citep[CRUDE,][]{crude} to that end. Fourth, we illustrate that the inverse variance weighing method, commonly used to combine predictions from members of an ensemble, improves predicted means but results in poorly calibrated uncertainties. Finally, we demonstrate, for the first time with a tabular dataset, the improvement in predicted negative log likelihood (NLL) resulting from \textit{robust}, domain-aware calibration \citep[][]{robust_calibration}. 

\begin{figure*}
    \centering
    \includegraphics[width=.98\linewidth]{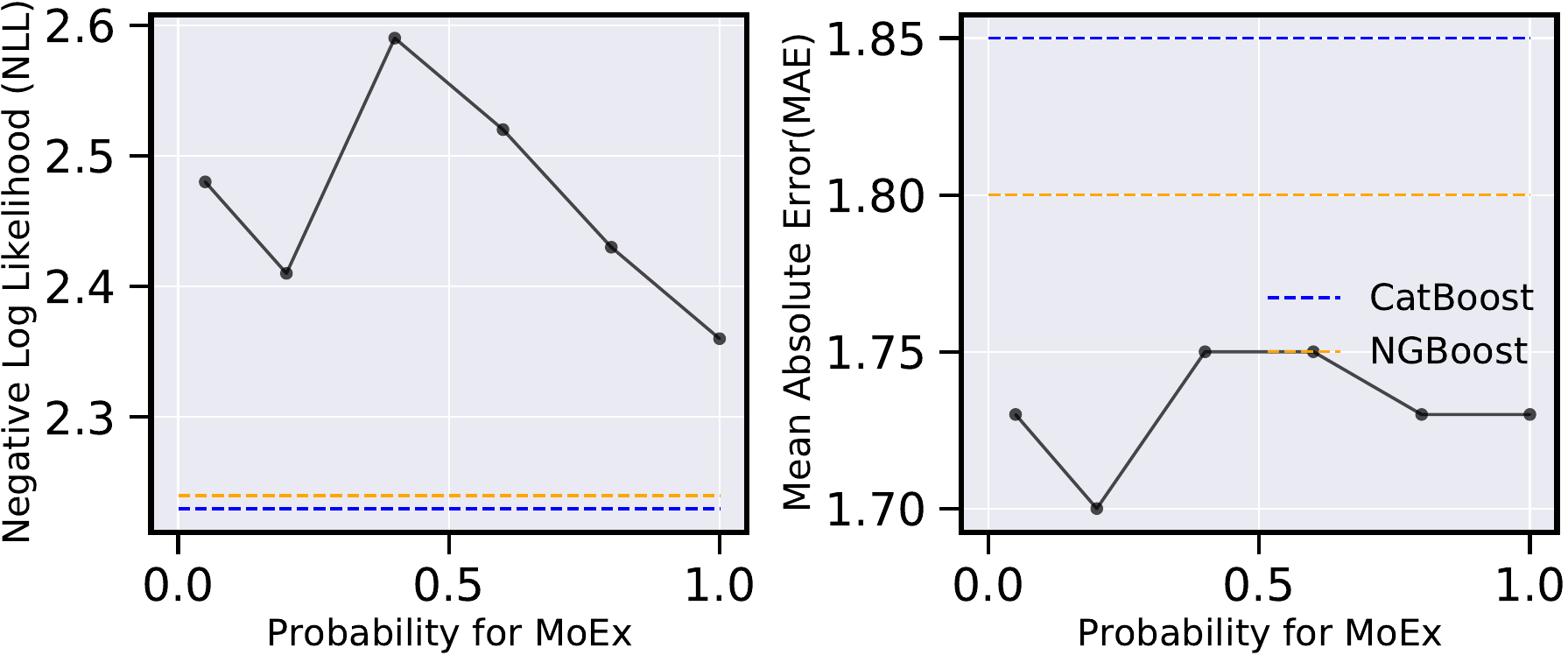}
    \caption{Impact of varying the probability parameter \textit{p} in MoEx. \textit{p} is the probability that a given sample will be augmented. \textbf{Left:} Negative log likelihood of \textit{robust, post-hoc} predictions on \texttt{TEST} vs. \textit{p} and mean absolute error. \textbf{Right:} Mean absolute error instead of NLL. For both plots, lower is better. We denote by dashed orange and blue lines the respective predicted metrics from \texttt{NGBoost} and \texttt{CatBoost} (see last lines in Tables \ref{table:results_ngb} and \ref{table:results_cb}, respectively).}
    \label{fig:mdn_moex_probability}
\end{figure*}

\section{Data}\label{sec:data}

We employ the \texttt{Weather Prediction} dataset developed by Yandex Research for the Shifts Challenge. This dataset comprises several meteorological features from three major global weather prediction models, as well as air temperatures 2 metres above ground (\texttt{fact\_temperature}), for a diverse set of latitudes and longitudes. The goal of this competition is to predict air temperature at 2 meters above ground, given all available weather station measurements and multiple weather forecast model predictions. While the training data has information about location on earth and the timestamp when predictions from the 3 global forecasting systems were made, these meta-data are assumed to be missing for the test set.

There are two specific set of feature categories that make up this dataset: weather-related features and meteorological features. Weather related features consist of sun evaluation at the current location, climate values of temperature, pressure and topography. Meteorological parameters are details based on pressure and surface level data from the weather prediction models.  There are five climate types associated with the data: Tropical, Dry, Mild Temperate, Snow and Polar. The data are split into \textit{training}, \textit{development} and \textit{evaluation} sets based on the climate and time of year, in order to provide a clear distributional shift in the weather data. The \textit{development} set is first sub-divided into ID-development and OoD-development (in-distribution an out of distribution, respectively), and is designed to simulate the \textit{evaluation} set. In this paper, we only use the first two datasets as these were the only ones released during the first phase of the competition. The training data consists of features with climate types as Tropical, Dry and Mild Temperature. This data is recorded from the $1^{st}$ of September 2018, to the $8^{th}$ of April, 2019. dev-ID has the same climate types and time of year (first and last thirds of the year) as the training data, whereas dev-OoD consists of samples only with the climate type Snow, and during the middle third of the year. The evaluation dataset, of course, has been shifted based on climate and time. It consists of features with Snow and Polar climate types, recorded from the $14^{th}$ May, 2019 to the $8^{th}$ of July,2019.\footnote{We mention this only for sake of exposition but remind the reader that we do not actually use the evaluation dataset in this work.} The training, dev-ID and dev-OoD datasets show a clear distributional shift, presenting a challenge to develop models robust to these shifts and competent at generalisation to OoD data in the real world. %We split the training data into six separate sets, based on the three types of climate (tropical, dry, and mild temperate), and the time of year (early and late); we expound upon our methodology in detail in the section below.

\iffalse
\begin{figure*}
    \centering
    \includegraphics[width=.98\linewidth]{figures/scatter_dfdevin_weekofyear_vs_climate.pdf}
    \caption{Six different subsets in the training data, based on \texttt{climate} and \texttt{week\_of\_year}.}
    \label{fig:split}
\end{figure*}
\fi

\section{Method}\label{sec:method}
%Regularization methods:
%\begin{enumerate}
%    \item Gradient Centralization
%    \item Adaptive Gradient Clipping
%    \item Sharpness Aware Minimization
%    \item Stochastic Weight Averaging
%    \item Fisher Penalty
%    \item Autoclip
%    \item Moment Exchange
%\end{enumerate}

%Neural Networks methods:
%\begin{enumerate}
    %\item Mixture Density Network (choi 2018 + bishop old paper)
    %\item Evidential Regression
%\end{enumerate}

\begin{figure}[!ht]
    \centering
    \includegraphics[width=.74\linewidth]{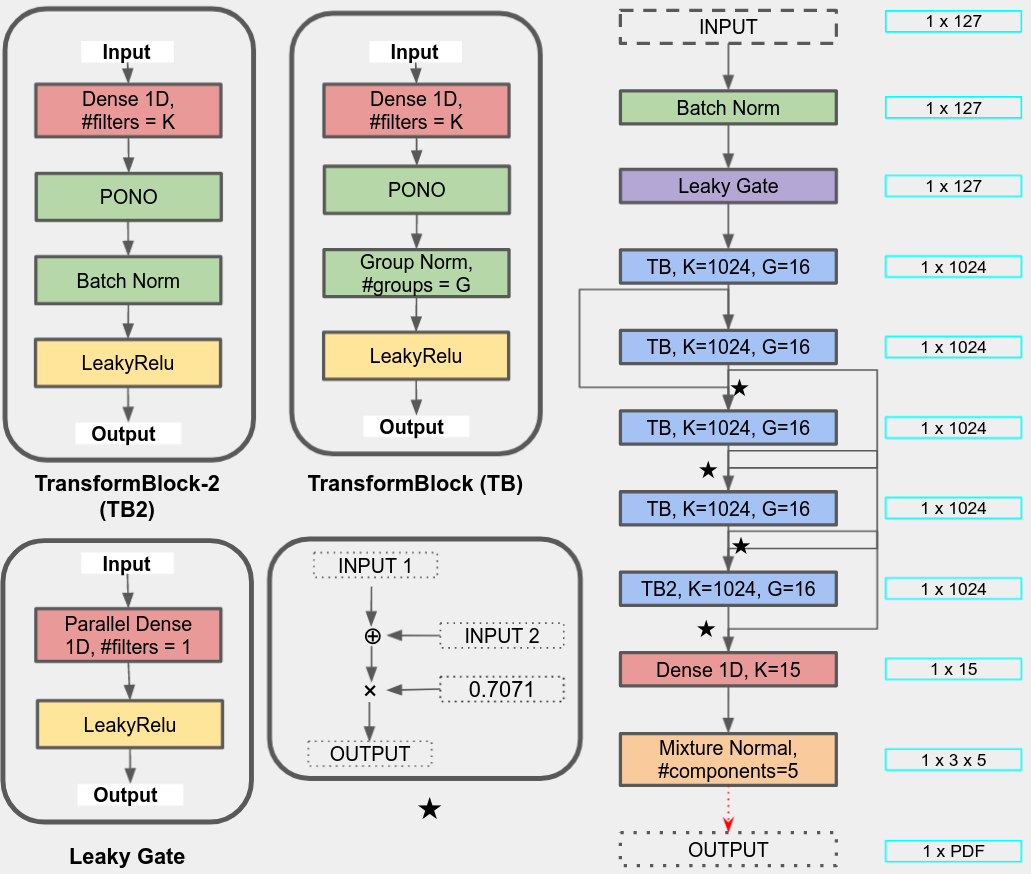}
    \caption{Architecture of the mixture density network \citep[MDN,][]{mdn0, mdn1}. PONO is the positional normalization layer \cite{pono}, and the use of LeakyGate is inspired by \cite{simple_mods_tabular}. We model the output variable \texttt{fact\_temperature} conditioned on the input variables using a mixture of 5 $\beta$ distributions.}
    \label{fig:mdn_architecture}
\end{figure}

\begin{enumerate}
        \item We divide the given training dataset $\texttt{T\_ALL}$ with $\sim$3.1 million samples into six disjoint datasets ($\texttt{T\_E}1$, $\texttt{T\_E}2$, $\texttt{T\_E}3$, $\texttt{T\_E}4$, $\texttt{T\_E}5$, and $\texttt{T\_E}6$) based on \texttt{climate} and \texttt{week\_of\_year}. We use the ID part of the development dataset as our validation data $\texttt{V\_ALL}$, and the OoD part as the test set $\texttt{TEST}$. We divide the former it into $\texttt{V\_E}1$, $\texttt{V\_E}2$, $\texttt{V\_E}3$, $\texttt{V\_E}4$, $\texttt{V\_E}5$, and $\texttt{V\_E}6$. The \texttt{E}s stand for environment, to highlight the fact that each of the six datasets come from a different domain/environment.  %, as shown in Figure \ref{fig:split}.
        From each of these 14 datasets (7 training and 7 validation), as well as from $\texttt{TEST}$, we drop the first six columns as these are to be treated as meta-data and not expected to be present in the final evaluation dataset. %Next, we combine both datasets, and the results for these are in Figure \ref{fig:onetone_d15}. We note that results using both datasets are more accurate, as can be deduced from the RMSE, MAE, and BE values in both these figures.
        \item We now have seven unique sets of training-validation splits upon which we we train separate models. We scale each of these separately. We standardize the input features (to mean of 0 and standard deviation of 1) of the training datasets ($\texttt{T\_E}1$ through $\texttt{T\_E}6$, and $\texttt{T\_ALL}$), and then re-scale them to lie between 0 and 1. We use the derived statistics to re-scale $\texttt{V\_E}1$ through $\texttt{V\_E}6$, and $\texttt{V\_ALL}$ as well as the respective copies of the test set $\texttt{TEST}$. We similarly normalize the six output columns of \texttt{fact\_temperature}, and then follow this up by min-max scaling them such that the training set values lie between 0.1 and 0.9. This is because of our models, the MDN (visualized in Figure \ref{fig:mdn_architecture}) uses a $\beta$-likelihood for the output variable, which requires it to lie between 0 and 1. We leave a little buffer zone of 0.1 on each size to allow for OoD predictions.
        %\item For modelling, we use an MDN \citep{mdn0, mdn1} for each of the six training sub-sets $\texttt{T\_E}1$ through $\texttt{T\_E}6$. All models follow the same architecture, shown in Figure \ref{fig:mdn_architecture}. %We choose the size and number of layers via trial-and-error, by training on $\texttt{T}$ and minimizing loss on the validation set $\texttt{VAL}$.
        \item We choose the hyperparameters for \texttt{NGBoost}\footnote{\texttt{n\_estimators}=500, \texttt{max\_depth}=12, \texttt{colsample\_by\_node}=0.3, \texttt{subsample}=0.8, \texttt{eta}=0.1, \texttt{num\_parallel\_trees}=3, \texttt{min\_child\_weight}=40, \texttt{gamma}=10, \texttt{reg\_lambda}=5, \texttt{reg\_alpha}=5, \texttt{distribution}=`normal'.}
        and \texttt{CatBoost}\footnote{\texttt{iterations}=500, \texttt{l2\_leaf\_reg}=10, \texttt{border\_count}=254, \texttt{depth}=10, \texttt{learning\_rate}=0.03, \texttt{use\_best\_model}=True, \texttt{loss}=`RMSEWithUncertainty'.} by trial-and-error. We use these two models as they allow to derive both aleatoric and epistemic uncertainties easily. For each type of model, we create 10 models with different seeds, and average the predictions from each such that the final mean on a test sample is the mean of the 10 predicted means, the final aleatoric uncertainty is the mean of the 10 predicted aleatoric uncertainties, and the final epistemic uncertainty is the variance of the 10 predicted means.
        \item For the MDN, we use 5 mixture components, negative log-likelihood (NLL) as the loss function, LAMB \citep{optimizer_lamb} as the optimizer, and a batch size of 512. We use a cosine decay learning rate scheduler \citep{cosine_decay_lr_scheduler} varying between $1e-3$ and $1e-4$, with 2 cycles each of length 15 epochs. We save and revert to model weights which result in the lowest loss values on the respective validation sets.
        \item For the MDN, we leverage moment exchange \citep[MoEx,][]{moex} to augment training data online. MoEx uses two hyperparameters: \textit{p}, the probability that a given sample will be augmented, and $\lambda$, the weight of a sample in a binary mixture of itself with another randomly selected sample (see original paper for details). We pick $p=0.20$ and draw lambda from a peaked $\beta$ distribution, such that it is more often than note close to 0.5 ($\lambda\sim\beta(100,100)$). In addition we use gradient centralization \citep[GC,][]{grad_cen} and stochastic weight averaging \citep[SWA,][]{swa} to smooth the loss landscape and improve generalization.
        %\item Each of the six MDNs above comprises of several regularization methods, shown to improve generalization. These include sharpness aware minimization \citep[SAM,][]{sam} with regularization constant \texttt{sam\_rho}, adaptive gradient clipping \citep[AGC,][]{adap_grad_clip} with regularization constant \texttt{clip\_thresh}, gradient centralization \citep[GC,][]{grad_cen}, stochastic weight averaging \citep[SWA,][]{swa}, and label distribution smoothing \citep[LDS,][]{deep_imb_reg}. We set \texttt{sam\_rho} and \texttt{clip\_thresh} to $0.15$ and $0.001$ respectively, via trial-and-error, similar to Step 3 above.
        \item Next, we combine the six sets of predictions from each of the individual training sets ($\texttt{T\_E}1$ through $\texttt{T\_E}6$) by inverse variance scaling.%, and conflation \citep{conflation}.
        \item We \textit{post-hoc} calibrate all predicted aleatoric uncertainties using CRUDE \citep{crude}, a state-of-the-art method for regression problems.
        \item When the training set is \texttt{T\_ALL}, we also calibrate the predicted aleatoric uncertainties on a per-domain basis \citep{robust_calibration}. Specifically, after training a model on \texttt{T\_ALL}, instead of using \texttt{V\_ALL} as the calibration set, we instead use $\texttt{V\_E}1$ through $\texttt{V\_E}6$ as individual calibration sets. For each of the six runs, we store the Gaussian NLL \citep{ibm_uncertainty_toolbox} derived from the calibrated aleatoric uncertainties on the respective validation/calibration sets, then find the $\texttt{V\_E}i$ with the minimum NLL -- this is the calibration set with respect to which \textit{post-hoc} calibration will result in maximum gain.
        \item Finally, to test the sensitivity of MDN with MoEx on the probability of augmentation \textit{p}, we run experiments with $p=0.05, 0.40, 0.60, 0.80, 1.00$; see Figure \ref{fig:mdn_moex_probability}.
        %\item We then train using the original training set $\texttt{TRAIN}$, use $\texttt{VAL}$ for validation, and predict results on $\texttt{TEST}$.
         %\item Feature engineering for time: NGBoost cannot, at least yet, handle time series data. To adequately leverage the temporal information in datasets \DDemo and \DSummary, we convert the column in each with time-stamps to three different features: hour-of-day, day-of-week, and week-of-year. All three range from 0 to 1. To account for the cyclic nature of these quantities, we then further convert them to a $\sin$ and a $\cos$ term each, thus converting the original single column with time-stamps to six separate columns. 
        %\item One-hot encoding of patient IDs: We have 37 patients, each with a unique patient ID. We one-hot encode these, thus converting the patient-ID column in both datasets \DDemo and \DSummary to 37 columns.
        %\item Split into training, validation, and test sets based on date: For each patient, we first order the the samples chronologically. Next we pick the first $70\%$ of the samples for training, the next $15\%$ for validation, and the last $15\%$ for testing. Finally, we collate the all data points for each patient into three bins for training, validation, and testing.
\end{enumerate}
For tracking our experiments and logging all metrics, we leverage Weights \& Biases \citep{wandb}\footnote{\url{https://wandb.ai/site}}.
%Gradient Boosted Decision Trees:
%\begin{enumerate}
%    \item NGBoost
%    \item Catboost
%    \item XGBoost
%    \item LGBM
%\end{enumerate}

\begin{table*}
  \caption{Relative performance of MDN+MoEx on the OoD test set ($\texttt{TEST}$), when trained individually on training datasets from the six environments ($\texttt{T\_E}1$ through $\texttt{T\_E}6$), when ensembling these results using inverse variance scaling, and when trianing on the combined dataset \texttt{T\_ALL}. The first two lines in each field correspond to raw and calibrated predictions. For the last cell with \texttt{T\_ALL}, the last line corresponds to robustly calibrated predictions. `M.T.' stands for mild temperate. For `Inverse Variance', the second row is not the results of calibration via CRUDE, but the inverse variance weighted means of the means and standard deviations from the calibrated predictions from the six domains (i.e., a weighted average of the second rows in the top six cells).}
  \label{table:results_mdnmoex}
  \centering
  \begin{tabular}{cccccccc}
    \toprule
    %\multicolumn{2}{c}{Part}                   \\
    %\cmidrule(r){1-2}
    Domain & Dataset & MAE ($\downarrow$) & RMSE ($\downarrow$) & BE ($|\downarrow|$) & IS ($|\downarrow|$) & ACE ($|\downarrow|$) & NLL ($\downarrow$)\\
    \otoprule
    \multirowcell{2}{M.T.\\Early} & \multirowcell{2}{$\texttt{T\_E}1$\\} & 2.32 & 3.11 & -0.40 & 0.04 & -0.23 & 3.00\\
                                   & & 2.32 & 3.11 & -0.42 & 0.05 & -0.29 & 3.46\\
    \midrule
    \multirowcell{2}{M.T.\\Late} & \multirowcell{2}{$\texttt{T\_E}2$\\} & 2.02 & 2.69 & -0.44 & 0.04 & -0.25 & 3.09\\
                                   & & 2.03 & 2.70 & -0.49 & 0.04 & -0.23 & 2.97\\
    \midrule
    \multirowcell{2}{Dry\\Early} & \multirowcell{2}{$\texttt{T\_E}3$\\} & 2.38 & 3.28 & 0.51 & 0.04 & -0.30 & 3.6\\
                                   & & 2.38 & 3.27 & 0.50 & 0.03 & -0.24 & 3.1\\
    \midrule
    \multirowcell{2}{Dry\\Late} & \multirowcell{2}{$\texttt{T\_E}4$\\} & 2.05 & 2.99 & -0.01 & 0.03 & -0.19 & 2.80\\
                                   & & 2.06 & 3.08 & 0.19 & 0.03 & -0.17 & 2.69\\
    \midrule
    \multirowcell{2}{Tropical\\Early} & \multirowcell{2}{$\texttt{T\_E}5$\\} & 3.06 & 3.90 & 1.87 & 0.06 & -0.21 & 3.47\\
                                   & & 3.09 & 3.93 & 1.93 & 0.07 & -0.30 & 4.84\\
    \midrule
    \multirowcell{2}{Tropical\\Late} & \multirowcell{2}{$\texttt{T\_E}6$\\} & 2.66 & 3.48 & 0.21 & 0.05 & -0.26 & 3.61\\
                                   & & 2.65 & 3.47 & 0.13 & 0.06 & -0.30 & 4.08\\
    \midrule
     & \multirowcell{2}{Inverse\\Variance} & 1.81 & 2.41 & 0.14 & 0.04 & -0.44 & 7.39\\
                                   & & 1.85 & 2.48 & 0.25 & 0.05 & -0.45 & 8.50\\
    \midrule
    \multirowcell{3}{All\\} & \multirowcell{3}{$\texttt{T\_ALL}$} & 1.74 & 2.33 & -0.15 & 0.03 & -0.08 & 2.26\\
                                 &  & 1.74 & 2.33 & -0.12 & 0.03 & -0.18 & 2.50\\
                                 &  & 1.73 & 2.33 & 0.12 & 0.03 & -0.17 & 2.48\\
    \bottomrule
  \end{tabular}
\end{table*}

\section{Results}\label{sec:results}
We compare all predictions using six metrics: mean absolute error (MAE), root mean squared error (RMSE), bias error (BE), interval sharpness \citep[IS,][]{gneiting_metrics, bracher_metrics, gilda_cfht, gilda_cfht_neurips}, average calibration error \citep[ACE][]{gneiting_metrics, bracher_metrics, gilda_cfht, gilda_cfht_neurips}, and Gaussian negative log likelihood \citep[NLL,][]{ibm_uncertainty_toolbox}. For experiments with MDN as the model, we show results in Table \ref{table:results_mdnmoex}. With \texttt{NGBoost} and \texttt{CatBoost} as the models, we show results in Tables \ref{table:results_ngb} and \ref{table:results_cb}, respectively. A few observations become apparent:
\begin{enumerate}
    \item The variance in metrics when trained on individual domains is quite high. In a real-life scenario of domain generalization, when it is difficult or even impossible to say which one of the training environments at hand might be the most similar to a given test environment, such high variance is undesirable.
    \item While inverse variance ensembling -- one of the most common methods of ensembling predictions -- might improve deterministic metrics (MAE, RMSE, MAE), they invariably deteriorate the probabilistic ones (ACE and NLL).
    \item In most cases, calibration improves predictions. In the handful of cases where it results in worse performance ($\texttt{T\_E}5$ and $\texttt{T\_E}6$ cells in Tables \ref{table:results_mdnmoex}, \ref{table:results_ngb} and \ref{table:results_cb}, and `inverse variance' cell in Table \ref{table:results_mdnmoex}), this can be attributed to the domain shift between the source and target domains. $\texttt{T\_E}5$ and $\texttt{T\_E}6$ datasets are from the tropical environment and are `far away', semantically, from \texttt{T\_TEST} which is in snowy climes. Consequently, \textit{post-hoc} calibrating predictions from datasets that are already poor representatives of a test set does more harm than good.
    \item From Table \ref{table:mdn_probability}, first rows of all cells, we see that the MDN's performance is mostly agnostic to the choice of the MoEx hyperparameter $p$ (judging by say MAE and NLL), except for high values of $p$ (see the $p=1.00$ cell). However, calibratiion -- even more so, robust calibration -- removes, to a large extent, the variance in performance, a desirable quality in domain generalization (see second and third rows of all cells). This strengthens further our case that both calibration, and when in a multi-environment setting, robust calibration, should be an indispensable tool in a researcher's toolkit when making probabilistic predictions.
    \item From Figure \ref{fig:mdn_moex_probability} we make two observations. First is that while (robustly calibrated) MAE with the MDN is lower than the (robustly calibrated) MAE from at all values of $p$, the opposite is true for NLL; we relegate further exploration of this to future work. Second, results based on the dataset under consideration suggest that $p=1$ is a good estimate for the MoEx hyperparameter, provided it that predictions are followed by robust calibration. 
\end{enumerate}

\section{Future Work}\label{sec:future_work}
This article represents only intermediate results based on the training and development sets available at the time of writing. As the immediate next step we will make predictions on the recently open-sourced evaluation dataset, and compare our results with those of the winners of the competition. We will also experiment with other neural network and boosted tree architectures, such as deep evidential regression \citep{der}, \texttt{XGBoost} \citep{xgboost}, and  \texttt{LightGBM} \citep{lgbm}. We will experiment with methods to ensemble predictions from different architectures; recent work has shown that such a hybrid ensemble can provide superior predictions than either machine learning- or deep learning-based models alone \citep{revisiting_dnn_tabular}. For a fairer comparison, we will also extensively optimize the hyperparameters of all models under consideration. In addition, we will investigate potential performance improvements by preprocessing the input data to increasing its SNR \citep{gilda_cont_norm_wavelet_kalman, gilda_cont_norm_wavelet_kalman_code} .Finally, we will experiment with other domains of training besides supervised, such as domain adaptation \citep{dann, gilda_sfh_domain_adaptation}, imbalanced risk minimization \citep[IRM,][]{irm}, and feature calibration \citep{taufe}.

\appendix

\bibliographystyle{plainnat}%{acm.bst}
\bibliography{main.bib}

\newpage
\section{Tables}\label{sec:tables}
\begin{table}[!ht]
\caption{Same as Table \ref{table:results_mdnmoex}, but using \texttt{NGBoost} as the predictor.}
  \label{table:results_ngb}
  \centering
  \begin{tabular}{cccccccc}
    \toprule
    %\multicolumn{2}{c}{Part}                   \\
    %\cmidrule(r){1-2}
    Domain & Dataset & MAE ($\downarrow$) & RMSE ($\downarrow$) & BE ($|\downarrow|$) & IS ($|\downarrow|$) & ACE ($|\downarrow|$) & NLL ($\downarrow$)\\
    \otoprule
    \multirowcell{2}{M.T. \\ Early} & \multirowcell{2}{$\texttt{T\_E}1$\\} & 2.09 & 2.77 & -0.90 & 0.03 & -0.06 & 2.35\\
                                   & & 2.01 & 2.68 & -0.66 & 0.03 & -0.00 & 2.31\\
    \midrule
    \multirowcell{2}{M.T. \\ Late} & \multirowcell{2}{$\texttt{T\_E}2$\\} & 1.92 & 2.57 & -0.62 & 0.03 & -0.06 & 2.29 \\
                                   & & 1.88 & 2.52 & -0.46 & 0.03 & -0.02 & 2.25 \\
    \midrule
    \multirowcell{2}{Dry \\ Early} & \multirowcell{2}{$\texttt{T\_E}3$\\} & 2.12 & 2.85 & -0.80 & 0.03 & -0.03 & 2.41\\
                                   & & 2.08 & 2.80 & -0.64 & 0.03 & 0.02 & 2.38\\
    \midrule
    \multirowcell{2}{Dry \\ Late} & \multirowcell{2}{$\texttt{T\_E}4$\\} & 1.96 & 2.61 & -0.38 & 0.03 & -0.03 & 2.31\\
                                   & & 1.93 & 2.58 & -0.20 & 0.03 & 0.01 & 2.29\\
    \midrule
    \multirowcell{2}{Tropical \\ Early} & \multirowcell{2}{$\texttt{T\_E}5$\\} & 4.31 & 5.56 & 3.57  & 0.11 & -0.17 & 2.89\\
                                   & & 4.51 & 5.82 & 3.84 & 0.11 & -0.16 & 2.91\\
    \midrule
    \multirowcell{2}{Tropical \\ Late} & \multirowcell{2}{$\texttt{T\_E}6$\\} & 4.44 & 5.65 & 3.72 & 0.11 & -0.08 & 2.89\\
                                   & & 4.60 & 5.85 & 3.92 & 0.11 & -0.02 & 2.91\\
    \midrule
     & \multirowcell{2}{Inverse\\Variance} & 1.85 & 2.49 & -0.20 & 0.04 & -0.35 & 3.72\\
                                   & & 1.84 & 2.48 & -0.04 & 0.04 & -0.32 & 3.41\\
    \midrule
    \multirowcell{3}{All\\} & \multirowcell{3}{$\texttt{T\_ALL}$} & 1.81 & 2.44 & -0.28 & 0.03 & -0.02 & 2.23\\
                                   & & 1.79 & 2.42 & -0.09 & 0.03 & 0.01 & 2.22\\
                                   & & 1.80 & 2.46 & 0.51 & 0.03 & 0.00 & 2.24\\
    \bottomrule
  \end{tabular}
\end{table}

\begin{table}[!ht]
\caption{Same as Table \ref{table:results_mdnmoex}, but using \texttt{CatBoost} as the predictor.}
  \label{table:results_cb}
  \centering
  \begin{tabular}{cccccccc}
    \toprule
    %\multicolumn{2}{c}{Part}                   \\
    %\cmidrule(r){1-2}
    Domain & Dataset & MAE ($\downarrow$) & RMSE ($\downarrow$) & BE ($|\downarrow|$) & IS ($|\downarrow|$) & ACE ($|\downarrow|$) & NLL ($\downarrow$)\\
    \otoprule
    \multirowcell{2}{M.T. \\ Early} & \multirowcell{2}{$\texttt{T\_E}1$\\} & 1.74 & 2.34 & -0.19 & 0.03 & -0.02 & 2.21\\
                                   & & 1.73 & 2.33 & -0.05 & 0.03 & -0.01 & 2.21\\
    \midrule
    \multirowcell{2}{M.T. \\ Late} & \multirowcell{2}{$\texttt{T\_E}2$\\} & 1.83 & 2.47 & -0.30 & 0.03 & -0.04 & 2.24\\
                                   & & 1.81 & 2.45 & -0.17 & 0.03 & -0.04 & 2.24\\
    \midrule
    \multirowcell{2}{Dry \\ Early} & \multirowcell{2}{$\texttt{T\_E}3$\\} & 1.80 & 2.40 & -0.19 & 0.03 & -0.03 & 2.27\\
                                   & & 1.79 & 2.39 & -0.12 & 0.03 & -0.01 & 2.25\\
    \midrule
    \multirowcell{2}{Dry \\ Late} & \multirowcell{2}{$\texttt{T\_E}4$\\} & 1.87 & 2.51 & -0.48 & 0.03 & -0.05 & 2.29\\
                                   & & 1.85 & 2.49 & -0.36 & 0.03 & -0.03 & 2.27\\
    \midrule
    \multirowcell{2}{Tropical \\ Early} & \multirowcell{2}{$\texttt{T\_E}5$\\} & 2.34 & 3.13 & 0.57 & 0.08 & -0.24 & 3.21\\
                                   & & 2.36 & 3.18 & 0.77 & 0.08 & -0.21 & 3.09 \\
    \midrule
    \multirowcell{2}{Tropical \\ Late} & \multirowcell{2}{$\texttt{T\_E}6$\\} & 2.22 & 2.97 & 0.28 & 0.07 & -0.22 & 3.04\\
                                   & & 2.24 & 3.01 & 0.54 & 0.07 & -0.20 & 2.99\\
    \midrule
     & \multirowcell{2}{Inverse\\Variance} & 1.83 & 2.45 & 0.01 & 0.05 & -0.41 & 5.59\\
                                   & & 1.82 & 2.44 & 0.16 & 0.05 & -0.40 & 5.30\\
    \midrule
    \multirowcell{3}{All\\} & \multirowcell{3}{$\texttt{T\_ALL}$} & 1.88 & 2.56 & -0.41 & 0.03 & -0.02 & 2.24\\
                                   & & 1.86 & 2.53 & -0.27 & 0.03 & -0.03 & 2.24\\
                                   & & 1.85 & 2.52 & -0.19 & 0.03 & -0.02 & 2.23\\
    \bottomrule
  \end{tabular}
\end{table}

\begin{table}[h!]
  \caption{Performance of the MDN with MoEx augmentation at different values of the probability parameter \textit{p}. We assume a $\lambda \sim0.5$ throughout. Training set is \texttt{T\_ALL}, validation set is \texttt{V\_ALL}. In each cell, the first row contains metrics from raw predictions on \texttt{TEST}, second row from CRUDE-calibrated predictions using \texttt{V\_ALL} as the calibration set, and the final row contains metrics when we leverage robust (per-domain) calibration as described in Section \ref{sec:method}.}
  \label{table:mdn_probability}
  \centering
  \begin{tabular}{ccccccc}
    \toprule
    %\multicolumn{2}{c}{Part}                   \\
    %\cmidrule(r){1-2}
    Probability (\textit{p}) & MAE ($\downarrow$) & RMSE ($\downarrow$) & BE ($|\downarrow|$) & IS ($|\downarrow|$) & ACE ($|\downarrow|$) & NLL ($\downarrow$)\\
    \otoprule
    \multirowcell{3}{0.05\\} & 1.74 & 2.33 & -0.15 & 0.03 & -0.08 & 2.26\\
                                   & 1.74 & 2.33 & -0.12 & 0.03 & -0.18 & 2.50\\
                                   & 1.73 & 2.33 & 0.12 & 0.03 & -0.17 & 2.48\\
    \midrule
    \multirowcell{3}{0.20\\} & 1.72 & 2.29 & -0.21 & 0.03 & -0.10 & 2.27\\
                                   & 1.71 & 2.28 & -0.11 & 0.03 & -0.14 & 2.37\\
                                   & 1.70 & 2.29 & 0.16 & 0.03 & -0.15 & 2.41\\
    \midrule
    \multirowcell{3}{0.40\\} & 1.78 & 2.36 & -0.32 & 0.03 & -0.13 & 2.40\\
                                   & 1.76 & 2.35 & -0.21 & 0.03 & -0.17 & 2.51\\
                                   & 1.75 & 2.34 & -0.12 & 0.03 & -0.19 & 2.59\\
    \midrule
    \multirowcell{3}{0.60\\} & 1.76 & 2.35 & -0.07 & 0.03 & -0.11 & 2.34\\
                                   & 1.76 & 2.35 & -0.12 & 0.03 & -0.16 & 2.45\\
                                   & 1.75 & 2.35 & 0.02 & 0.03 & -0.17 & 2.52\\
    \midrule
    \multirowcell{3}{0.80\\} & 1.73 & 2.33 & -0.07 & 0.02 & -0.10 & 2.30\\
                                   & 1.73 & 2.33 & 0.07 & 0.03 & -0.15 & 2.41\\
                                   & 1.73 & 2.34 & 0.10 & 0.03 & -0.15 & 2.43\\ 
    \midrule
    \multirowcell{3}{1.00\\} & 1.79 & 2.38 & -0.51 & 0.03 & -0.06 & 2.27\\
                                   & 1.75 & 2.35 & -0.31 & 0.03 & -0.13 & 2.37\\
                                   & 1.73 & 2.33 & -0.15 & 0.03 & -0.12 & 2.36\\
    \bottomrule
  \end{tabular}
\end{table}

\iffalse
%\newpage
\section{Figures}\label{sec:figures}
\begin{figure}[!ht]
    \centering
    \includegraphics[width=.74\linewidth]{figures/mdn_architecture.png}
    \caption{Architecture of the mixture density network \citep[MDN,][]{mdn0, mdn1}. PONO is the positional normalization layer \cite{pono}, and the use of LeakyGate is inspired by \cite{simple_mods_tabular}. We model the output variable \texttt{fact\_temperature} conditioned on the input variables using a mixture of 5 $\beta$ distributions.}
    \label{fig:mdn_architecture}
\end{figure}
\fi

\end{document}